\documentclass{article} % For LaTeX2e
\usepackage{iclr2023_conference,times}

\usepackage{amsmath,amsfonts,bm}

\def\eqref#1{equation~\ref{#1}}
\def\1{\bm{1}}

\DeclareMathAlphabet{\mathsfit}{\encodingdefault}{\sfdefault}{m}{sl}
\SetMathAlphabet{\mathsfit}{bold}{\encodingdefault}{\sfdefault}{bx}{n}

\usepackage{graphicx}
\usepackage{caption}
\usepackage{hyperref}
\usepackage{url}
\usepackage{array}
\usepackage{pifont}
\usepackage{booktabs}
\usepackage{tabularx}
\usepackage{adjustbox}
\usepackage{multirow}
\usepackage{makecell}
\usepackage{enumitem}
\usepackage{xspace}

\usepackage{tcolorbox}
\usepackage{booktabs,amsfonts,dcolumn}
\usepackage{hyperref}
\usepackage{url}
\usepackage{amsmath,amsthm,amsfonts,amssymb,bm,stmaryrd,bbm}
\usepackage{framed}
\usepackage{color}
\definecolor{shadecolor}{rgb}{0.92,0.92,0.92}
\definecolor{aqua}{rgb}{0.0, 1.0, 1.0}
\definecolor{azure(colorwheel)}{rgb}{0.0, 0.5, 1.0}
\definecolor{bleudefrance}{rgb}{0.19, 0.55, 0.91}
\definecolor{brandeisblue}{rgb}{0.0, 0.44, 1.0}
\definecolor{blue(ryb)}{rgb}{0.01, 0.28, 1.0}
\definecolor{lightblue}{RGB}{207,226,243}

\usepackage{cleveref}
\crefformat{section}{\S#2#1#3} % see manual of cleveref, section 8.2.1
\crefformat{subsection}{\S#2#1#3}
\crefformat{subsubsection}{\S#2#1#3}

\usepackage{algorithmic}
\usepackage[linesnumbered,ruled]{algorithm2e}
\SetKwRepeat{Do}{do}{while}%

\usepackage{cleveref}
\usepackage{wrapfig}
\usepackage{subcaption}
\usepackage{floatrow}

\newcommand{\tabincell}[2]{\begin{tabular}{@{}#1@{}}#2\end{tabular}}

\usepackage{pifont}

\newcommand{\ours}{\textsc{CoG}\xspace}

\usepackage{times}
\usepackage{latexsym}

\usepackage[T1]{fontenc}
\usepackage[utf8]{inputenc}

\usepackage{microtype}

\usepackage{inconsolata}

\usepackage{hyperref}
\usepackage{url}

\title{Copy is All You Need}

\iclrfinalcopy

\author{
 \textbf{Tian Lan}$^{\diamondsuit,\heartsuit,*}$%\thanks{Work done during an internship at Tencent AI Lab.}
 \quad
 \textbf{Deng Cai}$^{\diamondsuit,}$\thanks{~Contributed Equally.} $^{~,\dagger}$ \quad
 \textbf{Yan Wang}$^{\diamondsuit,}$\thanks{~Corresponding authors.} \quad
 \textbf{Heyan Huang}$^{\heartsuit}$  \quad
 \textbf{Xian-Ling Mao}$^{\heartsuit}$  \\
 $^{\diamondsuit}$Tencent AI Lab\\
 $^{\heartsuit}$School of Computer Science and Technology, Beijing Institute of Technology\\
 \texttt{\{lantiangmftby,thisisjcykcd,yanwang.branden\}@gmail.com}\\
 \texttt{\{hhy63,maoxl\}@bit.edu.cn}}

\begin{document}

\maketitle
\begin{abstract}
    The dominant text generation models compose the output by sequentially selecting words from a fixed vocabulary. In this paper, we formulate text generation as progressively copying text segments (e.g., words or phrases) from an existing text collection. We compute the contextualized representations of meaningful text segments and index them using efficient vector search toolkits. The task of text generation is then decomposed into a series of copy-and-paste operations: at each time step, we seek suitable text spans from the text collection rather than selecting from a standalone vocabulary. Experiments on the standard language modeling benchmark (WikiText-103) show that our approach achieves better generation quality according to both automatic and human evaluations. Besides, its inference efficiency is comparable to token-level autoregressive models thanks to the reduction of decoding steps. We also show that our approach allows for effective domain adaptation by simply switching to domain-specific text collection without extra training. Finally, we observe that our approach attains additional performance gains by simply scaling up to larger text collections, again without further training.\footnote{Our source codes are publicly available at \url{https://github.com/gmftbyGMFTBY/Copyisallyouneed}.}
\end{abstract}
    
    \section{Introduction}
    % traditional neural text generation task
    Most neural language models (LMs) process text generation tasks by making a series of next-token predictions in an autoregressive manner \citep{radford2019language,dai2019transformer,khandelwal2019generalization,Shi2022EffiditYA}. Specifically, LMs generate the next-token distribution over a fixed vocabulary for any given prefix. Then, the next token is selected by a chosen decoding method, such as greedy search and nucleus sampling \citep{holtzman2019curious}. This process continues until some stop condition is reached. For example, a special end-of-generation token is emitted, or the generated text reaches the maximum length limit.
    
    % our work
    Unlike traditional neural language models, we reformulate text generation by copying text segments from existing text collections. The text segments can be of variable lengths, including single words and multi-word phrases. For clarity, we will use the term ``phrase'' to refer to any contiguous text segments, and a single word can also be seen as a phrase of length 1. We compute a contextualized vector representation for each phrase and pack them into an offline index. At each decoding step, a suitable phrase is retrieved from the offline index and appended to the current prefix. In other words, the next-token predictions in traditional neural language models are replaced by a series of copy-and-paste operations.

    %advantages
    Our proposed model, named \textbf{\ours} (short for \textsc{\textbf{Co}py-\textbf{G}enerator}), enjoys the following advantages. First, our method selects \textit{phrases in specific contexts} rather than standalone tokens in a fixed vocabulary. It potentially allows for more accurate candidate representation and selection. Second, our method allows \textit{training-free adaptation} to new knowledge sources because the text collection can be updated in a plug-and-play fashion. It could benefit application scenarios such as domain adaptation and data expansion/filtering. Third, our method allows a sequence of multiple tokens (i.e., multi-word phrase) to be generated in one single step. It could reduce the total number of decoding steps, leading to improved inference efficiency.
    
    % experiment
    We conduct extensive experiments to verify the effectiveness of our proposed \ours. On the standard language modeling benchmark (WikiText-103), our proposed \ours substantially outperforms standard baselines on automatic metrics (26.14 vs. 23.43 MAUVE \citep{pillutla2021mauve}) and human evaluation (48\% vs. 28\% human preference). 
    Moreover, when we directly switch the text collection from the WikiText-103 corpus to a domain-specific corpus, Law-MT \citep{koehn2017six}, our proposed \ours outperforms strong baselines on this domain adaption setting (28.14 vs. 26.85 MAUVE and 52\% vs. 36\% human preference) without any domain-specific training.
    Furthermore, when we scale up the text collection of \ours to a larger one, the En-Wiki dataset, we obtain additional gain (26.97 vs. 23.43 MAUVE), again without any further training. 
    Our contributions can be summarized as follows:
    \begin{itemize}
        \item We propose \ours, a method that reformulates text generation tasks as a series of copy-and-paste operations from existing text collections.
        \item We show that \ours can outperform standard neural language model baselines on existing language modeling benchmarks.
        \item We demonstrate that \ours allows for training-free adaptations to larger text collections and domain-specific text collections.
    \end{itemize}
    
    \section{Background: Neural Text Generation}

    Neural text generation can be divided into two categories: (1) unconditional text generation; (2) conditional text generation. Unconditional text generation (or language modeling) aims to generate a coherent text continuation given a prefix. In this case, language models perform generation using a density estimation over sequences $p_{\theta}(x)$.
    Conditional text generation aims to generate text with some condition $c$ and instead estimates the probability of $p_{\theta}(x|c)$. Its typical applications include machine translation \citep{sutskever2014sequence,bahdanau2014neural}, summarization \citep{see2017get}.
    Throughout this paper, our discussion will be focused on unconditional text generation, however, our approach can be readily adapted to conditional text generation as well.

    The canonical approach to language modeling factors the generation in an autoregressive left-to-right manner $p_{\theta}(x_{0:n})=\prod_{i=1}^{n} p(x_i|x_{< i})$. In this case, text generation is reduced to the task of repeatedly predicting the next token conditioned on the partial sequence (i.e., prefix) generated so far $p(x_i|x_{< i})$. The model often consists of two parts: (1) a prefix encoder and (2) a set of token embeddings. The prefix encoder is often parameterized by the Transformer architecture \citep{vaswani2017attention}, which transforms any prefix into a fixed-sized vector representation $h_i \in \mathbb{R}^d = {\rm PrefixEncoder}(x_{< i}) $. Then, the probability of the next token being $w$ is calculated as
    \begin{equation}
    p_{\theta}(x_i=w|x_{< i}) = \frac{\exp(v_w \cdot h_i)}{\sum_{w\in V}\exp(v_w \cdot h_i)},
    \nonumber
    \end{equation}
    where $v_w$ is the context-independent token embedding representing the token $w$, and $V$ is the pre-defined vocabulary consisting of all possible tokens. Based on the chosen decoding method, such as greedy search and nucleus sampling \citep{holtzman2019curious}, the next token is selected according to the probability distribution over the fixed vocabulary $V$. This process is repeated in an autoregressive manner, until some stop condition is reached, e.g., the maximum length of generation is reached.

    \section{\textsc{Copy-Generator}}
        \begin{figure*}[t]
          \center{\includegraphics[width=1.0\textwidth]{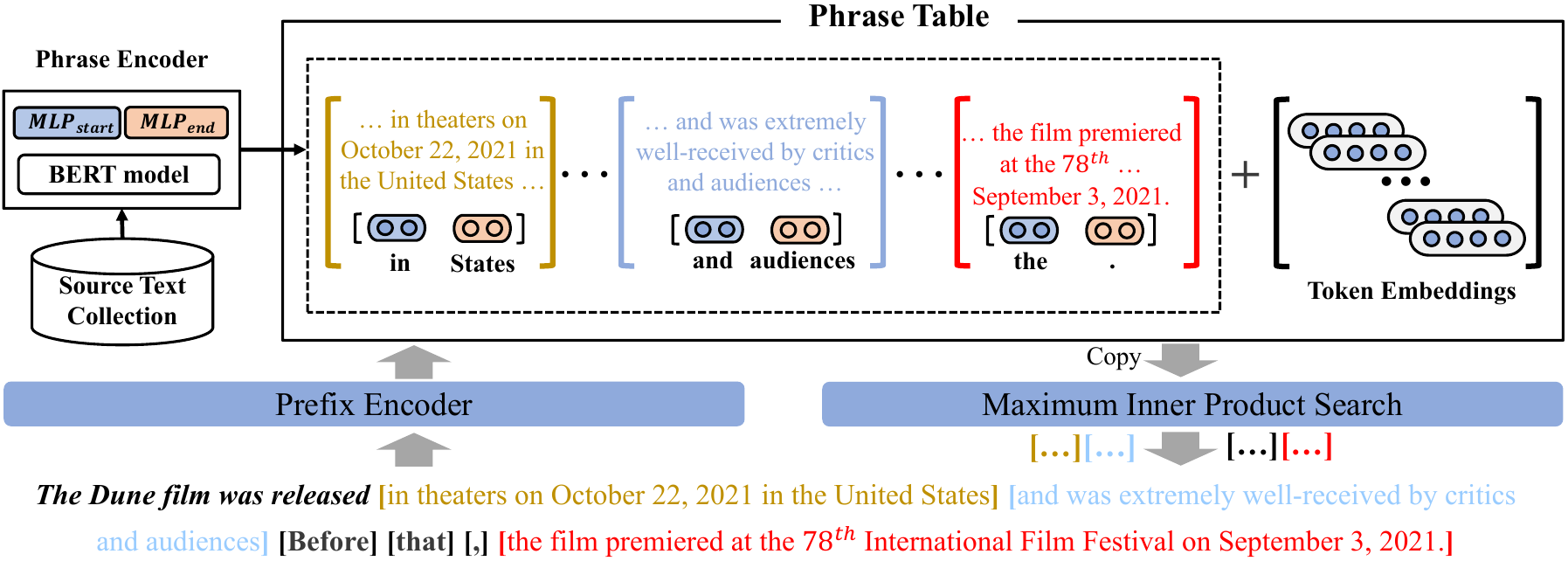}}
          \caption{The overview of our proposed \ours. Given the prefix \textit{The Dune film was released}, \ours retrieve 3 phrases (in different colors) from the documents and generates 3 tokens (\textit{Before}, \textit{that}, and the comma \textit{,}) from the fixed vocabulary to form the whole generation.}
          \label{img:overview}
        \end{figure*}
    Unlike traditional language models that compute the next token distribution over a fixed vocabulary that is usually composed of words or sub-words \citep{sennrich2016neural,kudo-richardson-2018-sentencepiece}, our proposed \ours has a dynamic ``vocabulary'' that is dependent on the available source text collections. Each item in the ``vocabulary'' corresponds to a text segment (termed as \textit{phrase} in this paper) in the source text collection. Importantly, all phrases are context-sensitive. That is, the same phrases in different contexts are considered to be different. The overall framework is depicted in Figure \ref{img:overview}.
    
    Formally, our approach assumes a set of source documents $\{D^1, \ldots, D^n\}$ is available. For each document $D^i$, a phrase $k= D^i_{s:e}$ of length $e-s+1$ can be extracted, where $s$ and $e$ mark the start and end positions of the phrase in the document, respectively. We denote all the phrases in the source text collection as $\mathcal{P}$. For a given prefix $x_{<i}$, we aim to select the best phrases that can form a coherent text continuation following the prefix. To this end, we compute a contextualized representation for each phrase $p_k \in \mathbb{R}^d = {\rm PhraseEncoder}(s, e, D^i)$ using a phrase encoder. Thus, a phrase table $\{(k, p_k) | k \in \mathcal{P}\}$ can be constructed. Similar to traditional language models, at test time, \ours also employs a prefix encoder to map the prefix $x_{<i}$ into a vector representation $q_i$. The fitness of a phrase $k$ to the prefix $x_{<i}$ is then measured by the dot product of their vector representations $p_k$ and $q_i$:
        \begin{equation}
          p(k|x_{<i}) \propto \exp(p_k \cdot q_i).
         \label{eq:fitness}
        \end{equation}
    At each time step, a suitable phrase is selected and appended to the current prefix accordingly.

    Note that the size of the phrase table can be up to billions.  To search over this large candidate pool, we pre-compute the phrase representations and use a coarse-to-fine search pipeline based on maximum inner product search (MIPS) \citep{johnson2019billion}. The details are deferred to Section \ref{sec:impl_details}. Moreover, to support the scenarios where no suitable phrases are available, we also add the context-independent token embeddings $\{(w, v_w) | w \in V\}$ in standard LMs to the phrase table.
    \paragraph{Ethical Consideration}
    The text generated by \ours contains text segments copied from other documents, which may cause copyright disputes in real-world applications. Therefore, there are a few things to be considered: (1) The copyright of the source text documents needs to be carefully checked. One should not use documents with strict copyright protection and/or private information; (2) It is recommended to quote the original source explicitly, especially when the retrieved phrases are long.

    \subsection{Model Architecture}
    \label{sec:model_architecture}
    As illustrated in Figure \ref{img:overview}, our proposed model consists of three major components:
    (1) a \textit{prefix encoder} that maps prefixes to fixed-sized representations;
    (2) a context-dependent \textit{phrase encoder} that computes the vector representations of the phrases in the source text collection;
    (3) a set of context-independent \textit{token embeddings} similar to the one used in standard neural language models.

    \paragraph{Prefix Encoder}
    The prefix encoder is responsible for encoding the prefix $x_{<i}$ into a vector representation for the next-phrase prediction. We treat the prefix as a sequence of tokens (previously predicted phrases are split into tokens as well) and encode them using the standard Transformer architecture with causal attention \citep{vaswani2017attention,radford2019language}. Causal attention only allows each position in the input sequence to attend to its preceding positions. Therefore, the prefix representation can be computed incrementally as the generation progresses, leading to faster inference. Concretely, the prefix encoder transforms a prefix $x_{< i}$ of length $i$ into a matrix $\mathcal{H}_i \in \mathbb{R}^{i \times dL}$, where $d$ is the hidden dimension and $L$ is the number of Transformer layers. The computation can be written as: 
        \begin{equation}
          \begin{split}
            \mathcal{H}_{i+1}={\rm PrefixEncoder}(x_{i},  \mathcal{H}_{i}).
          \end{split}
          \nonumber
        \end{equation}
    We use the hidden state of the last token as the prefix representation $q_i$.

    \paragraph{Phrase Encoder}
    Given a set of source documents $\{D^1,..., D^n\}$, the phrase encoder computes the vector representations of all the phrases in the documents. 
    % LT add
    % Note that the prefix could also be treated as a source document, allowing us to generate coherent text by copying the phrases in the prefix.
    Inspired by previous work~\citep{lee2016learning,seo2018phrase,lee2021learning}, we construct context-dependent phrase representations as follows. For a document $D=D_1, \ldots, D_m$ of length $m$, we first apply a deep bidirectional Transformer \citep{devlin2018bert} to obtain contextualized token representations $\mathcal{D}\in \mathbb{R}^{m\times d_t}$, where $d_t$ is the dimension of token representations. Then, we apply two MLPs models, MLP$_{\texttt{start}}$ and MLP$_{\texttt{end}}$, to convert $\mathcal{D}$ into start and end token representations $\mathcal{D}_{\texttt{start}}, \mathcal{D}_{\texttt{end}} \in \mathbb{R}^{m \times \frac{d}{2}}$, respectively:
    \begin{equation}
     \begin{split}
    \mathcal{D}_{\texttt{start}}={\rm MLP}_{\texttt{start}}(\mathcal{D}), \mathcal{D}_{\texttt{end}}={\rm MLP}_{\texttt{end}}(\mathcal{D}).
    \end{split}
    \nonumber
    \end{equation}
    For each phrase $D_{s:e}$ that starts at $s$ and ends at $e$ in the document, we use the concatenation of the corresponding start and end vectors as the phrase representation.
        \begin{equation}
        %\small
          \begin{split}
            {\rm PhraseEncoder}(s, e, D)=[\mathcal{D}_{\texttt{start}}[s];\mathcal{D}_{\texttt{end}}[e]] \in \mathbb{R}^{d}
          \end{split}
         %\nonumber
         \label{eq:extract_phrase_rep}
        \end{equation}
     The advantages of the above representation method are that (1) we only need to encode the document once to obtain all phrase representations; and (2) we only need to store all the token representations instead of all phrase representations.
    \paragraph{Context-Independent Token Embeddings}
    Although \ours can copy phrases from other documents, we would like to retain the generalization capability to compose output with standalone tokens. This can be especially useful when there is no suitable phrase in the source text collection. Therefore, we also add the traditional context-independent token embeddings  $\mathcal{V}\in \mathbb{R}^{|V|\times d}$ to our phrase table. These tokens can be seen as phrases of length 1 without any context information. 
        
    \subsection{Model Training}
    %%%%%%%%%%%
    \ours decomposes the task of text generation into a series of copy-and-paste operations: at each time step, it selects the next phrase either from the source text collection or the fixed token vocabulary. In other words, phrases are used as the basic building blocks for text generation. To train \ours, each document in the training set is chunked into a sequence of phrases in a similar spirit. Specifically, we propose a greedy segmentation algorithm based on forward maximum matching. Taking a document $D=D_1, \ldots, D_{m}$ of $m$ tokens as an example, our algorithm segments the document from left to right. The first $i$ tokens will be cut out as a phrase if it can be found as a sub-sequence in other documents and $i$ is the maximum valid value. The above process is repeated until all tokens are cut out. Note that some resultant phrases can be single tokens in the fixed token vocabulary when no proper matching can be found. Detailed explanations of the phrase segmentation algorithm can be found in Appendix \ref{appendix:alg}.

    % LT revise 
    % TODO: meetup for the duplicate of the phrases before the softmax (去重和不去重对效果影响不大)，ICLR rebuttle提到了这个，我们这里也说明一下？
    Suppose that a document $D$ has been split into $n$ phrases $D=p_1, \ldots, p_n$. If the $k$-th phrase $p_k$ is copied from another document, let $D^k$ be the source document and let $s_k, e_k$ be the start and end positions of $p_k$ in $D^k$, the phrase encoder is used to extract its context-dependent phrase representations ${\rm PhraseEncoder}(s_k, e_k, D^k)$ (Eq. \ref{eq:extract_phrase_rep}). On the other hand, we directly retrieve the context-independent token embedding of $p_k$ if it is copied from the fixed token vocabulary. As illustrated by Eq. \ref{eq:fitness}, \ours relies on a shared vector space of prefix and phrase representations, where the representations of semantically coherent prefixes and phrases should be closer to each other while others should be pushed apart. We define the training loss for next-phrase predictions by using the InfoNCE loss with in-batch negatives \citep{karpukhin2020dense}:
        \begin{equation}
          \begin{split}
            &\mathcal{L}_{p}=-\frac{1}{n}\sum_{k=1}^n\log\frac{\exp(q_{k} \cdot p_{k})}{\sum_{p \in \mathcal{P}_k} \exp(q_{k} \cdot p_p) + \sum_{w \in V  } \exp(q_{k} \cdot v_w)}
          \end{split}
         \nonumber
        \end{equation}
        where $\mathcal{P}_k$ consists of all the phrases in the source document $D^k$,
        $V$ is the set of all tokens in the token vocabulary, and $q_{k}$ denotes the representation of the prefix preceding the phrase $p_{k}$ in $D$.

        Additionally, to retain the capability of token-level generation, we also train \ours with the standard token-level autoregressive loss.
        \begin{equation}
          \begin{split}
            & \mathcal{L}_t=-\frac{1}{m}\sum_{i=1}^{m}\log\frac{\exp(q_{i}, v_{D_i})}{\sum_{w \in V} \exp(q_{i}, v_w)}
          \end{split}
           \nonumber
        \end{equation}
        where $q_i$ denotes the prefix representation preceding the token $D_i$ in $D$. Finally, the training loss is the sum of these two losses:
        \begin{equation}
          \begin{split}
            & \mathcal{L}=\mathcal{L}_{p}+\mathcal{L}_t
          \end{split}
         \nonumber
        \end{equation}

    \section{Experimental Setup}
    \subsection{Baselines}
    We compare \ours with the following three baselines:
    \begin{itemize}%[wide=0.\parindent,noitemsep,topsep=0.em]
    \item \textbf{Transformer} \citep{vaswani2017attention} has been the \textit{de facto} model for neural language models. Concretely, we fine-tune the pre-trained GPT2 model \citep{radford2019language} in our experiments. 
    %which contains 12 transformer layers with hidden size of 768.
    \item \textbf{$k$NN-LM} \citep{khandelwal2019generalization} is a retrieval-augmented generation model, which extends a pre-trained neural language model by linearly interpolating its next token distribution with a $k$-nearest neighbors ($k$NN) model.%Concretely, the fine-tuned Transformer baseline is used to generate all the key-values pairs.
    \item \textbf{RETRO} \citep{borgeaud2022improving}\footnote{\url{https://github.com/lucidrains/RETRO-pytorch}.} is another retrieval-augmented generation model which combines a frozen BERT retriever, a differentiable encoder and a chunked cross-attention mechanism to predict next tokens. Since there is no pre-trained RETRO model that could be accessed, we train it from scratch on the WikiText-103 dataset. % with more training steps.
    \end{itemize}

    \subsection{Implementation Details}
    \label{sec:impl_details}
    All the baselines and our source codes are based on the popular Huggingface transformers package \citep{wolf-etal-2020-transformers}. 
    For a fair comparison, the prefix encoders in Transformer, $k$NN-LM, and \ours use the same model architecture as the pre-trained GPT2 model (12 layers, 12 heads, and 768 hidden dimensions) \citep{radford2019language}. 
    For the phrase encoder in \ours, we fine-tune the pre-trained BERT-base-cased model \citep{devlin2018bert} (12 layers, 12 heads, and 768 hidden dimensions). We train baselines and \ours for 400,000 steps on 8 Tesla-V100 GPUs.
    For all the baselines, the learning rate, dropout rate, and gradient clipping are set as 5e-5, 0.1, and 1.0, respectively.
    Due to memory limitation, the batch size is set to contain 256 phrases. For the BERT model in the phrase encoder, the maximum sequence length is set as 256. For the GPT2 model in the prefix encoder, the maximum sequence length is set as 512. Our proposed \ours contains overall 248M parameters from BERT and GPT2 models, and other baselines contain over 124M parameters.
    As suggested by \cite{borgeaud2022improving}, the hyper-parameters $\lambda$ and $\alpha$ of $k$NN-LM are set as 0.118 and 0.00785, respectively.
    
    To improve the inference efficiency of \ours, we encode all the documents in the source text collections offline. Note that retrieving from such a super large phrase collection faces severe challenges on the engineering side. This paper uses a coarse-to-fine pipeline to address this challenge. Specifically, we first use a document retriever to retrieve top-$k$ related documents for each given prefix. Then, their corresponding phrase representations are collected for selection. In this paper, a popular semantic matching model, DPR \citep{karpukhin2020dense} and a vector search toolkit, FAISS \citep{johnson2019billion} are used as the document retriever, which can recall documents that have similar topics with the prefix. The value $k$ is empirically set to 1024.

    \ours can be used with both greedy search and nucleus sampling. For greedy search, \ours selects the phrase that has the highest fitness score at each time step. As for nucleus sampling, we first obtain the next-phrase distribution by using the \texttt{softmax} function over the fitness scores of all candidate phrases. Then, the next phrase is sampled over this distribution.
    
    More details of the implementation can be found in Appendix \ref{appendix:dataset} and \ref{appendix:impl}.
    
    \subsection{Automatic Evaluation Metrics}
    For each document in the test set, we use the first 32 tokens as the prefix. The baselines and our proposed \ours generate text continuations of length 128 based on the same prefix. Following conventions \citep{welleck2019neural,Su2022ACF}, we use \textbf{greedy search} and \textbf{nucleus sampling} \citep{holtzman2019curious} ($p=0.95$) throughout our experiments.
    Following previous work \citep{welleck2019neural,Su2022ACF} and report the results on the following evaluation metrics:
    \begin{itemize}%[wide=0.\parindent,noitemsep,topsep=0.em]
        \item \textbf{MAUVE} \citep{pillutla2021mauve}, an efficient, interpretable, practical automatic evaluation, is highly coherent with human judgments and widely used to evaluate modern text generation models \citep{Su2022ACF,krishna2022rankgen}. In this paper, MAUVE leverages the GPT2-large model to generate the scores, and the scaling factor is set as 2.0.
        \item \textbf{Rep-$n$} \citep{welleck2019neural} measures the sequence-level repetition as the portion of duplicate n-grams in the generated text \citep{welleck2019neural}. For a generation text $x$, Rep-$n$ can be formulated as: $100\times (1.0-\frac{\rm |unique\ n-gram(x)|}{\rm |total\  n-gram(x)|})$. Higher Rep-$n$ denotes the severe degeneration problem in generations.
        \item \textbf{Diversity} \citep{welleck2019neural} measures the diversity of the generations, which is formulated as $\Pi_{n=2}^4(1-\frac{{\rm Rep}-n}{100}))$. Generations that have higher Diversity scores usually are more informative.
        %\item \textbf{Coherence} score \citep{Su2022AnES,Su2022ACF,Lan2022MomentumDO} automatically measures the semantic coherence between the prefix text $\boldsymbol{x}$ and the generated text $\hat{\boldsymbol{x}}$ using a massively pre-trained OPT-2.7B LM~\cite{Zhang2022OPTOP}, i.e. $\mathcal{M}$. Formally, the metric is defined as the averaged log-likelihood of the generated text conditioned on the prefix text as:
        % \begin{equation}
        % \nonumber
        % %\small
        %     \begin{split}
        %         \frac{1}{|\hat{\boldsymbol{x}}|} \sum_{i=1}^{|\hat{\boldsymbol{x}}|} \log p_{\mathcal{M}}\left(\hat{\boldsymbol{x}}_i \mid\left[\boldsymbol{x}: \hat{\boldsymbol{x}}_{<i}\right]\right),
        %     \end{split}
        %     \label{eq:coherence}
        % \end{equation}
        % where $[:]$ is the concatenation operation.
    \end{itemize}
    
    Note that previous work \citep{khandelwal2019generalization,dai2019transformer} often uses perplexity as the primary evaluation metric to measure the performance of language modeling. However, since our proposed \ours does not calculate next-token distributions over a fixed vocabulary, the comparison of perplexities is not reliable and thus omitted. However, we can test the perplexity of generated text using an external language model, and the results are shown in Appendix \ref{appendix:ppl}.
    
    \section{Experimental Results}

    In this paper, we evaluate baselines and our proposed \ours in three different settings:
    (1) standard language modeling; (2) domain adaption; (3) enlarged phrase index.

    \subsection{Language Modelling on WikiText-103}
    \label{sec:wikitext103}
    
    In this setting, models are trained on the training set of the WikiText-103 dataset and evaluated on its test set. 
    The WikiText-103 dataset \citep{merity2016pointer} contains an extensive collection of Wikipedia articles with over 100 million words, which is widely used to evaluate the performance of universal language modeling \citep{khandelwal2019generalization,dai2019transformer,Su2022ACF}.

    \begin{table*}[h]
        %\small
            \begin{center}
            \renewcommand{\arraystretch}{1.2}
            \setlength{\tabcolsep}{3.2pt}
            \scalebox{0.95}{
            
                \begin{tabular}{cccccccc}
\hline
\textbf{Model}                                                                                  & \textbf{Decoding} & \textbf{MAUVE}$\uparrow$& \textbf{Rep-2}$\downarrow$ & \textbf{Rep-3}$\downarrow$ & \textbf{Rep-4} $\downarrow$&\textbf{Diversity}$\uparrow$&\textbf{Latency (s)}$\downarrow$\\ \hline
\multirow{2}{*}{\textbf{Transformer}}      & greedy    & 19.87   & 43.56    &  38.55   &   35.5 &22.37    & 1.32      \\ 
  & nucleus &    23.43     &  5.10                &  1.33              &  0.50          & 93.22& 1.48     \\ \hline
\multirow{2}{*}{\textbf{$k$NN-LM}}  & greedy    & 19.92     & 43.79              &  38.76     &   35.69 &22.13&  10.36    \\
& nucleus &  22.50  &   \textbf{3.33}    &  \textbf{0.69}    &\textbf{0.21}   & \textbf{95.8}& 10.42     \\ \hline
\multirow{2}{*}{\textbf{RETRO}}  & greedy  &   21.19                 &  44.65      &   39.63     & 36.6  & 21.19&  4.39  \\
& nucleus &   22.86       &    6.21          &    1.93           & 0.86    & 91.19& 4.51    \\ \hline
\multirow{2}{*}{\textbf{\ours}}       & greedy    &  26.01      &    28.14    &      23.80     &  21.40 &  43.03&  \textbf{1.29}         \\ 
& nucleus &       \textbf{26.14}      &  7.31              &    2.66           & 1.28 &    89.07& 1.54       \\ \hline
\end{tabular}
            }
            \caption{The automatic evaluation on the test set of WikiText-103. As for each model with nucleus sampling, we run 10 times and recorded the average MAUVE and Diversity scores.
            }
            \label{tab:wikitext103}
            \end{center}
        \end{table*}
    
    \paragraph{Results}
    Table \ref{tab:wikitext103} shows the performance comparison between the baselines and our proposed \ours on the test set of the WikiText-103 corpus. It can be found that our proposed \ours substantially outperforms the Transformer and $k$NN-LM baselines on most metrics.
    Specifically, \ours improves MAUVE score over the best baseline (Transformer with nucleus sampling) from 23.43 to 26.14 -- an improvement of 2.71\%.
    Interestingly, although it is well known that greedy search could raise severe degeneration problems \citep{welleck2019neural}, 
    \ours with greedy search still outperforms the standard Transformer baseline with nucleus sampling, with 2.58\% improvements on MAUVE.
    This observation demonstrates that \ours is more robust and less prone to the degeneration problem, which can be considered as an additional bonus.
    %Moreover, the coherence of \ours significantly outperforms strong baselines when the nucleus sampling is conducted. This observation indicates that \ours could generate more coherent generations strong baselines.
    
    \begin{wraptable}[7]{R}{0.6\textwidth}
    %\begin{table}
    \vspace{-1em}
    %\small
    \centering
        \renewcommand{\arraystretch}{1.2}
        \setlength{\tabcolsep}{3.2pt}
        \scalebox{0.85}{
            \begin{tabular}{ccccccccc}
            \hline
            \textbf{Method} & \textbf{Uni-gram} & \textbf{2-gram} & \textbf{3-gram} & \textbf{4-gram} & \textbf{5-gram} & \textbf{6-gram} \\ \hline
            \textbf{Greedy}       & 0.583                  &      0.195           &  0.121               &   0.056              & 0.029                & 0.017              \\
            \textbf{Nucleus}       & 0.434                  &  0.219               & 0.181                & 0.09                &  0.048               &0.028            \\ \hline
            \end{tabular}
        }
        \caption{The statistics on the length of the copied phrases (on the test set of WikiText-103).}
        \label{tab:ngram_length_stat}
    \end{wraptable}
    \paragraph{Inference Speed}  Furthermore, we also compare the average time cost of different methods for completing the generation on the test set. Since the phrase representations in \ours are pre-computed offline, its encoding time cost is not included. The results are reported in Table \ref{tab:wikitext103}. As seen, \ours still achieves comparable inference efficiency with the standard Transformer baseline. The reason is that the copied phrases usually contain multiple tokens (the statistics of phrase length are shown in Table \ref{tab:ngram_length_stat}). As a result, \ours uses fewer decoding steps when generating the text of the same length.
    Unlike \ours that uses a coarse-to-fine search pipeline, $k$NN-LM conducts large-scale vector search at every decoding step. Its inference latency is much higher than Transformer, and \ours, which is aligned with previous work\citep{alon2022neuro}.
    %We count the distribution of the length of the copied phrases, which is shown in Table \ref{tab:ngram_length_stat}. It can be found that over 50\% of decoding steps contain phrases that have more than 2 tokens for \ours with nucleus sampling. Moreover, the average length and variance of these copied phrases are 2.18 and 1.82, respectively.
    \begin{wraptable}[7]{R}{0.45\textwidth}
    \vspace{-1em}
    \small
    \centering
        \renewcommand{\arraystretch}{1.2}
        \setlength{\tabcolsep}{3.2pt}
        \scalebox{1.0}{
            \begin{tabular}{cccc}
            \hline
            \textbf{Comparison} & \textbf{Better} & \textbf{No Prefer.} & \textbf{Worse} \\ \hline
            \tabincell{c}{\textbf{\ours vs.}\\\textbf{Transformer}}     & \textbf{48\%}                         & 24\%                        & 28\%                         \\\hline
            \end{tabular}
        }
        \caption{Human evaluation on the WikiText-103 corpus.}
        \label{tab:human_evaluation_wikitext103}
    \end{wraptable}
    \paragraph{Human Evaluation} To ensure the reliability of our evaluations, we also run human evaluation with three native-speaker graders from a third-party grading platform. Specifically, we randomly select 100 test prompts. For each test prompt, the annotators are given two continuations, in random order, which are generated by \ours and Transformer respectively. The annotators are asked to decide which one is better by considering the following aspects:
    \begin{itemize}[wide=0.\parindent,noitemsep,topsep=0.em]
        \item \textbf{Fluency}: Whether the generated text is fluent and easy to understand.
        \item \textbf{Informativeness}: Whether the generated text is diverse and contains interesting content. 
    \end{itemize}
    When annotators make different decisions on the same sample, we ask them to have a discussion and make the final decision. As shown in Table \ref{tab:human_evaluation_wikitext103}, our proposed \ours model significantly outperforms strong Transformer baseline, indicating its better generation quality.

\begin{figure*}[t]     
          \center{\includegraphics[width=1.0\textwidth] {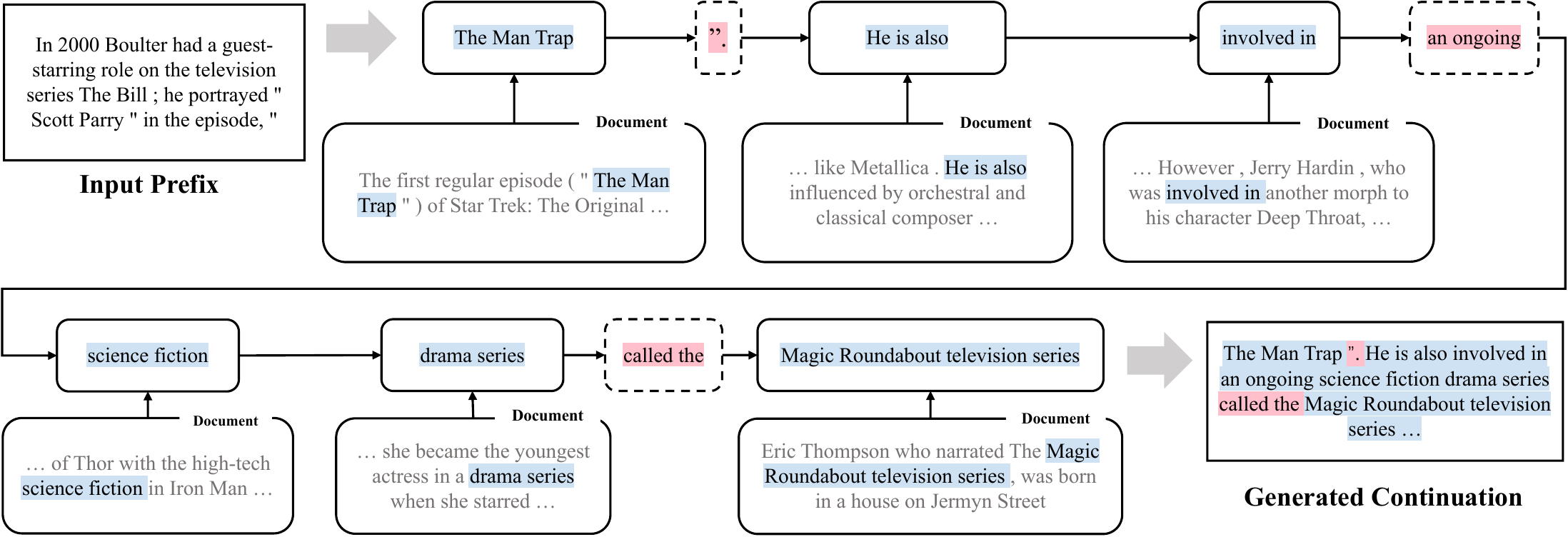}}
          \caption{An example generated by \ours on the test set of WikiText-103. The dotted squares denote that the content (highlighted in \colorbox{pink}{red})is copied from the token vocabulary, and the solid squares denote that the content (highlighted in \colorbox{lightblue}{blue}) is copied from other documents.}
          \label{img:case_study}
        \end{figure*}

    \paragraph{Case Study}
        For a better understanding of the performance of \ours, we present an example of the text continuations generated by our proposed \ours in Figure \ref{img:case_study}. It can be found that \ours can retrieve phrases that are semantically coherent and fluent for given prefixes.
        For example, at the second decoding step, \ours generate the punctuations $[$\textit{''}, \textit{.}$]$ from the pre-defined vocabulary to close the film name \textit{``The Man Trap''} and the sentence.
        Besides, at the ninth decoding step, \ours directly copied the named entity \textit{Magic Roundabout television series} from the related document. More examples can be found in Appendix \ref{appendix:case}.
        %On the other hand, we find that entity mismatch is a prevalent problem, likely because our retriever does not rely on lexical overlapping. We leave the investigation of more "accurate" document retrieval as future work.
    
    \subsection{Domain Adaption on Law-MT}
        In the domain adaption setting, the models trained on the WikiText-103 dataset are tested on a specific domain. Following previous work \citep{he2021efficient,alon2022neuro}, we use the English part of Law-MT \citep{koehn2017six}, which is an English-German translation dataset for law documents. The memory of $k$NN-LM, RETRO and \ours are constructed from the training set of Law-MT. We also present the performance of Transformer baselines with or without further fine-tuning on the training set of Law-MT.

            \begin{wraptable}[12]{R}{0.5\textwidth}
            \vspace{-1em}
    \small
    \centering
    \renewcommand{\arraystretch}{1.2}
    \setlength{\tabcolsep}{3.2pt}
    \scalebox{0.8}{
            \begin{tabular}{cccc}
\hline
\textbf{Model}                                                                                  & \textbf{Decoding}         & \textbf{MAUVE} $\uparrow$&\textbf{Diversity} $\uparrow$\\ \hline
\multirow{2}{*}{\textbf{Transformer w/o FT}}      & greedy    & 20.32&	70.66    \\ 
  & nucleus &  25.21&	93.88     \\ \hline
\multirow{2}{*}{\textbf{Transformer w/ FT}}      & greedy    & 23.00&	80.52    \\ 
  & nucleus &  26.85&	90.14     \\ \hline
\multirow{2}{*}{\textbf{$k$NN-LM}}      & greedy    & 23.31&	19.85 \\ 
  & nucleus & 24.75&	\textbf{94.60}    \\ \hline
\multirow{2}{*}{\textbf{RETRO}}      & greedy    & 18.70&71.14   \\ 
  & nucleus & 20.35&	94.81  \\ \hline
\multirow{2}{*}{\textbf{\ours}}      & greedy    & 21.31	&	84.32 \\ 
  & nucleus &  \textbf{28.14}&	92.56  \\ \hline
\end{tabular}      
    }
    \caption{The automatic evaluation on Law-MT. 
    %The memory of $k$NN-LM, RETRO, and \ours are built on the train set of Law-MT.
    }
    \label{tab:setup_2}
    \end{wraptable}
        \paragraph{Results}
        % KNNLM
        \vspace{-1em}
        As shown in Table \ref{tab:setup_2}, it can be observed that \ours even outperforms the Transformer model further fine-tuned on the Law-MT corpus (Transformer w/ FT). Specifically, \ours outperforms Transformer w/ FT by 2.93\% MAUVE score. The results indicate that \ours allows a single model to be specialized in different domains, by simply switching the source text collection. Although $k$NN-LM brings in higher Diversity scores, \ours surpasses it by 3.39\% MAUVE score, which shows \ours has higher generation quality in general.

    \begin{wraptable}[6]{R}{0.5\textwidth}
    \vspace{-1em}
\small
\centering
        \renewcommand{\arraystretch}{1.2}
        \setlength{\tabcolsep}{3.2pt}
        \scalebox{1.0}{
            \begin{tabular}{cccc}
            \hline
            \textbf{Comparison} & \textbf{Better} & \textbf{No Prefer.} & \textbf{Worse} \\ \hline
            \tabincell{c}{\textbf{\ours vs.}\\\textbf{Transformer w/ FT}}     & \textbf{52\%}                         & 12\%                        & 36\%                         \\\hline
            \end{tabular}
        }
        \caption{Human evaluation on Law-MT.}
    \label{tab:human_evaluation_lawmt}
\end{wraptable}
    \paragraph{Human Evaluation}
    We also conduct the human evaluation on the Law-MT corpus, which has a similar setup to that in (\cref{sec:wikitext103}). Table \ref{tab:human_evaluation_lawmt} shows that most of \ours's generations are better than a strong Transformer baseline. This observation demonstrates that \ours can even outperform the fine-tuned Transformer baseline without any domain-specific training.

    \subsection{Enlarged Phrase Index with En-Wiki}
    In the enlarged phrase index setting, we make use of a large text collection, the En-Wiki corpus, and test baselines on the test set of WikiText-103. The En-Wiki corpus contains a large-scale collection of Wikipedia articles with over 3 billion words, whose size is much larger than the WikiText-103 dataset. The memory of $k$NN-LM, RETRO, and \ours are built from the training set of En-Wiki\footnote{
    Due to the hardware limitation, RETRO uses the subset of the En-Wiki corpus (over 6 million chunks).}. 
    Similar to the domain adaption setting, we also present the results of Transformer baselines with or without further fine-tuning on the En-Wiki corpus.

        \paragraph{Results}
        
        The experimental results are shown in Table \ref{tab:setup_3}. 
        %\ours surpasses the Transformer baseline with and without further fine-tuning on the En-Wiki corpus by 3.54\% MAUVE scores. This is especially remarkable because \ours does not require any additional training, suggesting we can train \ours with a smaller corpus but leverage additional information in a larger corpus in a plug-and-play fashion.
        %\ours surpasses the Transformer baseline by 3.54\% MAUVE scores. 
        \ours with En-Wiki memory surpasses other strong baselines and \ours with WikiText-103 memory.
        This is especially remarkable because \ours does not require any additional training, suggesting we can train \ours with a smaller corpus but leverage additional information in a larger corpus in a plug-and-play fashion.
        Similar to the domain adaption setting, we also notice that, although $k$NN-LM baseline improves Diversity scores, it obtains a much lower MAUVE score than \ours (23.39 vs. 26.97).
        Note that the Transformer w/ FT is slightly worse than that without fine-tuning on the En-Wiki dataset. This phenomenon is mainly because there are deviations between En-Wiki and WikiText-103 datasets.

        \paragraph{Effects of Index Size}
        To further investigate how the size of the phrase index affects the generation quality, we randomly sample several subsets of the En-Wiki dataset with proportions from $0.1\%$ to $100\%$. As shown in Figure \ref{fig:intro-example}, when the proportion is less than $1\%$, \textbf{\ours} exhibits a similar quality, which is unsurprising since few enlarged documents are added to the phrase index. In contrast, once the proportion is larger than $1\%$, the larger the phrase index becomes, the better generation quality the model achieves. 

            \begin{minipage}{\textwidth}
        \begin{minipage}[h]{0.48\textwidth}
                \centering
    \renewcommand{\arraystretch}{1.2}
    \setlength{\tabcolsep}{3.2pt}
    \scalebox{0.8}{
                \begin{tabular}{cccc}
\hline
\textbf{Model}                                                                                  & \textbf{Decoding}         & \textbf{MAUVE} $\uparrow$&\textbf{Diversity} $\uparrow$\\ \hline
\multirow{2}{*}{\textbf{Transformer w/o FT}}      & greedy    &19.87&22.37  \\ 
  & nucleus & 23.43&  93.22   \\ \hline
\multirow{2}{*}{\textbf{Transformer w/ FT}}      & greedy    & 20.21&19.62  \\ 
  & nucleus & 21.31& 92.92   \\ \hline
%\multirow{2}{*}{\textbf{Transformer w/ FT}}      & greedy    & 20.00&20.29 \\ 
%  & nucleus & &    \\ \hline
\multirow{2}{*}{\textbf{$k$NN-LM}}      & greedy    &23.21 &20.33\\ 
  & nucleus &23.39 & \textbf{96.37}  \\ \hline
\multirow{2}{*}{\textbf{RETRO}}      & greedy    &19.75&21.15  \\ 
  & nucleus &22.87&91.09 \\ \hline
\multirow{2}{*}{\textbf{\ours}}      & greedy    & 24.68&40.45 \\ 
  & nucleus & \textbf{26.97} & 90.00\\ \hline
\end{tabular}
        
    }
    %\caption{The automatic evaluation on the test set of WikiText-103, the retrieval resources are built on the train set of En-Wiki. We runs 10 times and recorded the average scores. Transformer w/ FT and Transformer w/o FT denote the Transformer baseline with and without further fine-tuning on the train set of En-Wiki, respectively.}
    \makeatletter\def\@captype{table}\makeatother\caption{The automatic evaluation on the test set of WikiText-103, 
    the memory is built on the train set of En-Wiki. %We runs 10 times and recorded the average scores.
    Transformer w/ FT and Transformer w/o FT denote the Transformer baseline with and without further fine-tuning on the train set of En-Wiki, respectively.}
    \label{tab:setup_3}
        \end{minipage}
        \hspace{1em}
        \begin{minipage}[h]{0.48\textwidth}
        \vspace{-0.8em}
            \centering
            \includegraphics[width=1.0\textwidth] {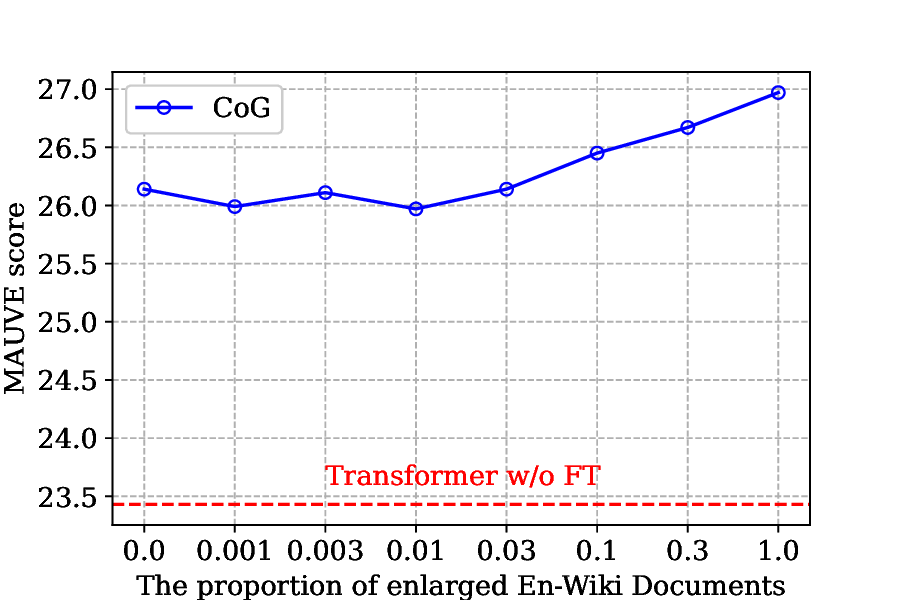}
            %\caption{Generation quality of \textbf{\ours} with different sizes of the phrase index. For each proportion (point in the X-axis), we sample 10 times and record the averaged MAUVE score. A proportion of $0.0$ indicates that only documents from WikiText-103 are used.}
            \makeatletter\def\@captype{figure}\makeatother\caption{Generation quality of \textbf{\ours} with different sizes of the phrase index. For each proportion (point in the X-axis), we sample 10 times and record the averaged MAUVE score. A proportion of $0.0$ indicates that only documents from WikiText-103 are used.}
            \label{fig:intro-example}  
        \end{minipage}
    \end{minipage}
        
        % In this part, we further analyze how the proportion of enlarged memory affects the generation quality. Specifically, we randomly sample a subset of En-Wiki documents with a proportion from $0.1\%$ to $100\%$. As shown in Figure \ref{fig:intro-example}, 
        % it can be observed that, before $1\%$ proportion, the generation quality is stable, since few En-Wiki documents are added to the offline index. Then, with the enlarging of document proportion ( from $1\%$ to $100\%$), the MAUVE score becomes consistently bigger, indicating a better generation quality. 
    \section{Related Work}
    
    \paragraph{Dense Retrieval}
    The dense retrieval technique \citep{karpukhin2020dense} has been widely used in many downstream NLP tasks, such as open-domain question answering \citep{karpukhin2020dense,lee2021learning}, open-domain dialogue systems \citep{lan2021exploring} and machine translation \citep{cai-etal-2021-neural}. Different from the traditional sparse retrieval system, such as BM25 and TF-IDF \citep{Robertson2009ThePR}, dense retrieval learns a shared vector space for queries and documents, where relevant pairs of query and document have smaller distances (i.e., higher similarities) than the irrelevant pairs.

    The most closely related work to our study is DensePhrase \citep{lee2021learning}. DensePhrase reformulates the question-answering task as a phrase retrieval problem, where phrases are directly retrieved and returned as answers to factual questions. Differently, our work aims to generate coherent text continuations through multiple rounds of phrase retrieval. Since the connection between two adjacent phrases should be coherent and fluent in the text generation task, it is much more difficult.
    
    \paragraph{Retrieval-Augmented Text Generation (RAG)}
    Retrieval-augmented text generation has gained increasing interest recently. Most prior work improves the generation quality (e.g., informativeness) of language models by grounding the generation on a set of retrieved materials (e.g., relevant documents) \citep{Li2022ASO,guu2020realm,hashimoto2018retrieve,Weston2018RetrieveAR,cai2019skeleton,cai-etal-2019-retrieval,khandelwal2019generalization,wu2019response,guu2020realm,lewis2020retrieval,borgeaud2022improving,2304.04487}. Our work is on this line of research but takes a radical step forward. Unlike prior work that builds the combinations of retrieval and generation, retrieval is generation in \ours.

    One contemporary work to our work is \cite{min2022nonparametric}, which shares the idea of replacing the fixed vocabulary with a nonparametric phrase table. However, \cite{min2022nonparametric} focuses on masked language modeling while our focus is on causal language modeling and text generation.
    
\section{Conclusion}
    In this paper, we reformulated text generation as progressively copying phrases from the massive text collection. Following this formalization, we proposed a novel neural text generation model, named \ours, which generates text by retrieving semantically coherent and fluent phrases from other documents. Experimental results proved the advantages of \ours over the strong baselines on three experimental settings: standard language modeling (WikiText-103), domain adaptation (Law-MT), and enlarged phrase index (En-Wiki).

% \section*{Limitations}

% While \ours achieves strong performance, it needs to copy phrases from the pre-processed phrases index. However, it is a huge engineering challenge for us to index all the phrases in the large-scale corpus.
% In this paper, we simply employ the recall-then-rerank pipeline to address this problem, i.e., we first recall a batch of documents related to the given prefix, and all the phrases are extracted from these documents online.
% Fortunately, we notice that this problem could be easily addressed by using some tricks proposed by previous works, for example, the DensePhrase \cite{lee2021learning}.
% In summary, it needs extra effort to build the overall phrase index from the large-scale corpus for \ours.

\section*{Justification of Changes}
Note that the experimental results in the current version have some changes from the previous version that has been reviewed. We made a number of revisions to the experiments according to the valuable suggestions from the reviewers.

\section*{Acknowledgement}
The authors thank the anonymous reviewers for their valuable suggestions and comments on our paper, which significantly improves the quality of our paper.

\bibliography{iclr2023_conference}
\bibliographystyle{iclr2023_conference}

\appendix

%\section{Appendix}
\section{Dataset Statistics}
\label{appendix:dataset}
The experiments in this paper include three benchmarks:
(1) WikiText-103; (2) English part of Law-MT; (3) En-Wiki.
The statistics of these benchmarks are shown in Table \ref{tab:benchmark_statistics}. En-Wiki corpus is used for the enlarged phrase index settings in this paper, containing over 4,848,348 long English Wikipedia documents.

\begin{table}[h]
\begin{tabular}{cccc}
\hline
\textbf{Benchmarks}   & \textbf{Train} & \textbf{Dev} & \textbf{Test} \\ \hline
\textbf{WikiText-103} & 1,801,350        & 3,760         & 4,358          \\
\textbf{Law-MT}       & 389,292         & 2,000         & 2,000          \\ \hline
\end{tabular}
\caption{The number of sentences in the WikiText-103 and Law-MT datasets.}
\label{tab:benchmark_statistics}
\end{table}

% \subsection{Details of Human Evaluation}
% \label{appendix:human_annotation}
% %Three native-speaker graders from the third-party grading platform are hired to annotate the text generated by \ours and baselines by considering the following aspects:
% %\begin{itemize}
% %    \item \textbf{Fluency}: Whether the generated text is fluent and easy to understand.
% %    \item \textbf{Informativeness}: Whether the generated text is diverse and contains interesting content. 
% %\end{itemize}
% Note that the overall annotation process is fully anonymous and will not raise the risk to their private information.
% All the graders are provided adequate payment for the annotation task in this paper. 
% Moreover, all the graders have authorized the human annotation data for our academic research.
% All the annotators are explained how the data would be used and their private data are not included during annotations (e.g., sexual orientation or political views under GDPR).

    \section{More Implementation Details}
    \label{appendix:impl}

    During training, the dynamic vocabulary of \ours contains two parts: (1) word-level vocabulary size (50257 in GPT2 vocabulary); (2) the phrases in a batch of training documents. During inference, the dynamic vocabulary consists of the word-level vocabulary and the phrases extracted from the Top-$k$ retrieved documents ($k$=1024 in this paper). The size of the pre-defined word-level vocabulary contains 50257 subwords. Since there are only a few documents encoded to extract the phrase representations, the average number of the phrase representations is 950,942.4 in the WikiText-103 test set when $K=1024$.

    %\ours can be used with both greedy search and nucleus sampling. For greedy search, \ours selects the phrase that has the highest fitness score at each time step. As for nucleus sampling, we first obtain the next-phrase distribution by using the \texttt{softmax} function over the fitness scores of all candidate phrases. Then, the next phrase is sampled over this distribution.

  \begin{wraptable}[10]{R}{0.45\textwidth}
  \vspace{-2em}
            \begin{center}
            \renewcommand{\arraystretch}{1.0}
            \setlength{\tabcolsep}{5pt}
            \scalebox{0.9}{
\begin{tabular}{ccc}
\hline
\multirow{2}{*}{\textbf{Models}} & \multicolumn{2}{c}{\textbf{Perplexity}} \\ \cline{2-3}
                                  & \textbf{greedy}    & \textbf{nucleus}   \\ \hline
\textbf{Transformer} &  3.26  &  37.11     \\
\textbf{$k$NN-LM}    & 3.48    &  78.01       \\
\textbf{RETRO}       &  3.27      &  36.40                  \\
\textbf{\ours}       & 10.41      &  27.24     \\\hline
\textbf{Ground-Truth}&\multicolumn{2}{c}{18.64}   \\\hline     
\end{tabular}
            }
            \caption{The perplexity on the test set of WikiText-103.}
            \label{tab:appendix_ppl}
            \end{center}
        \end{wraptable}
\section{Perplexity of Generated Text}
\label{appendix:ppl}
We calculate the perplexity of the generated texts under a large pre-trained language model (GPT2-Large). As shown in Table \ref{tab:appendix_ppl}, it can be found texts generated by greedy search can achieve very low perplexity scores (even much lower than the ground-truth)\footnote{Note that the original perplexity of GPT2-Large model on the test set of WikiText-103 is 22.05 \citep{radford2019language}. The gap between it and our results is caused by the different number of samples. In this study, we only use samples that have more than 32 tokens to generate text.}. This is expected as greedy search targets at likelihood maximization. Sampling-based decoding methods give much higher perplexity scores. Moreover, it is worth noting that \ours achieves the closest perplexity score to ground-truth. %This observation demonstrates that \ours's generation is much more similar to the human-written text, due to lots of copied phrase

\section{The Phrase Segmentation Algorithm}
	\label{appendix:alg}
	\ours takes phrases as the minimum units that can be put together to form a coherent document. To train \ours, we design a phrase segmentation algorithm to split each document in the training set into a sequence of phrases. This algorithm makes use of a forward maximum matching strategy to identify phrases. Maximum matching is one of the most popular structural segmentation algorithms. This method favors long phrases and is a greedy algorithm by design. Specifically, we treat each document as a sequence of tokens and scan each document from left to right. At each step, we search for the longest prefix of the unsegmented part that is also a sub-sequence of other documents other than the current document. If the length of that prefix is bigger than 2, we take that prefix as the next phrase. Otherwise, we take the first token as the next phrase and it is labeled as coming from the fixed token vocabulary. In both cases, we process the rest part of the current document recurrently. The algorithm can be very time-consuming because exhaustive searches over millions of documents are compute-intensive. Therefore, we propose an efficient approximation as follows. First, we retrieve the top-$k$ most similar documents for each document using the popular DPR model \citep{karpukhin2020dense}\footnote{\url{https://huggingface.co/facebook/dpr-ctx_encoder-single-nq-base}.}, and vector search toolkits, FAISS \citep{johnson2019billion}. Then, the phrase search only runs on the corresponding top-$k$ documents. The relevant documents usually have similar topics to the current document. The value of $k$ is set as 1024 in our experiments. The details of our proposed phrase segmentation algorithm can be found in Algorithm \ref{alg:psa}: SearchPhrase is a function that searches the cached token sequence (i.e., the current candidate for the next phrase) among the most relevant documents. It returns a label that denotes whether the phrase can be found and its position in the relevant documents.
	
	\IncMargin{1em}
	\begin{algorithm*}
	
	    \SetKwFunction{SearchPhrase}{SearchPhrase}
	
        \KwData{Document set: $\mathcal{D}=\{d_i, \{d_j\}_{j=1}^K\}_{i=1}^N$, where $d_i$ denotes the $i$-th document. $K$ denotes the number of retrieved documents. $N$ denotes the number of documents in the training set. The pre-defined maximum and minimum phrase lengths are $L_{max}$ and $L_{min}$.}
        \KwResult{Segmented document set by phrase granularity: $\mathcal{D}'=\{\{(p_{i,x},(d_j,{\rm pos_j})) \}_{x=1}^{||d_i||_p}\}_{i=1}^N$, where $p_{i,x}$ denotes the $x$-th phrase in $d_i$ that also appears in another document $d_j$ in position $j$. $||d_i||_p$ denotes the number of the collected phrases in $d_i$.}
        
        \BlankLine
        
        \textbf{Preprocess}: 
        split each document into token-level pieces by using the off-the-shelf tokenizer. 
        The preprocessed document set can be formulated as $\mathcal{D}=\{\{t_{i,x}\}_{x=1}^{||d_i||_t}, \{d_j\}_{j=1}^K\}_{i=1}^N$, where $t_{i,x}$ is the $x$-th token of $d_i$, which consists of $||d_i||_t$ tokens. Prepare the empty list $\mathcal{D}'=\{\}$, empty phrase cache cache$_p$=$\{\}$, and cached search success label label$_{last}$.\\
        
        \BlankLine
        
        \For{$i\leftarrow 1$ \KwTo $N$}{
            cursor=0\\
            PhraseCollection=$\{\}$\\
            \While{cursor$\leq ||d_i||_t$}{
                \eIf{$L_{min}$ $\leq$ len(cache$_p$) $\leq$ $L_{max}$}{
                    label$_{now}$, rest=SearchPhrase(cache$_p$)\\
                }{
                    \If{len(cache$_p$) $>$ $L_{max}$}{
                        cache$_{p}$=$\{\}$\\
                    }
                } 
                \eIf{label$_{last}$ is True and label$_{now}$ is False}{
                    cursor -= 1\\
                    PhraseCollection.append(cache$_p$, rest)\\
                    cache$_p$=$\{\}$
                }{
                   \If{label$_{last}$ is False and label$_{now}$ is False}{
                        PhraseCollection.append( cache$_p$, None)\\
                        cache$_p$=$\{\}$
                    }
                }
                cursor += 1\\
                label$_{now}$=label$_{last}$
            }
            $\mathcal{D}'$.append(${\rm PhraseCollection}$)
        }
        \caption{Phrase Segmentation Algorithm}
        \label{alg:psa}
    \end{algorithm*}
    \DecMargin{1em}

% \subsection{How the Number of Retrieved Documents affects the generation quality}
%     \label{appendix:size_documents}
%         \begin{figure}[h]     
%           \center{\includegraphics[width=\textwidth] {img/diversity_docnum.eps}}
%           \caption{How the number of the retrieved documents influences the diversity of the text continuations generated by our proposed \ours.}
%           \label{img:diversity_docnum}
%         \end{figure}
        
%         In this subsection, we analyze how the number of the top-$k$ documents influences the diversity of the text continuations generated by our proposed \ours. As shown in Figure \ref{img:diversity_docnum}, we could make the following conclusions: 
%         (1) with the number of retrieved documents increasing, the diversity of the text continuations becomes higher, no matter what decoding method is chosen;
%         (2) \ours with greedy search outperforms the Transformer model with nucleus sampling (0.848 vs. 0.83 Diversity score) when the number of retrieved documents is larger than 20. Noted that the parameters of nucleus sampling for the Transformer baseline are carefully selected by the grid search;
%         (3) \ours with nucleus sampling achieves comparable diversity metric (0.942 vs. 0.94) when the number of the documents is larger than 90.

\section{More Cases}
    \label{appendix:case}
\begin{figure*}[h]     
          \center{\includegraphics[width=\textwidth] {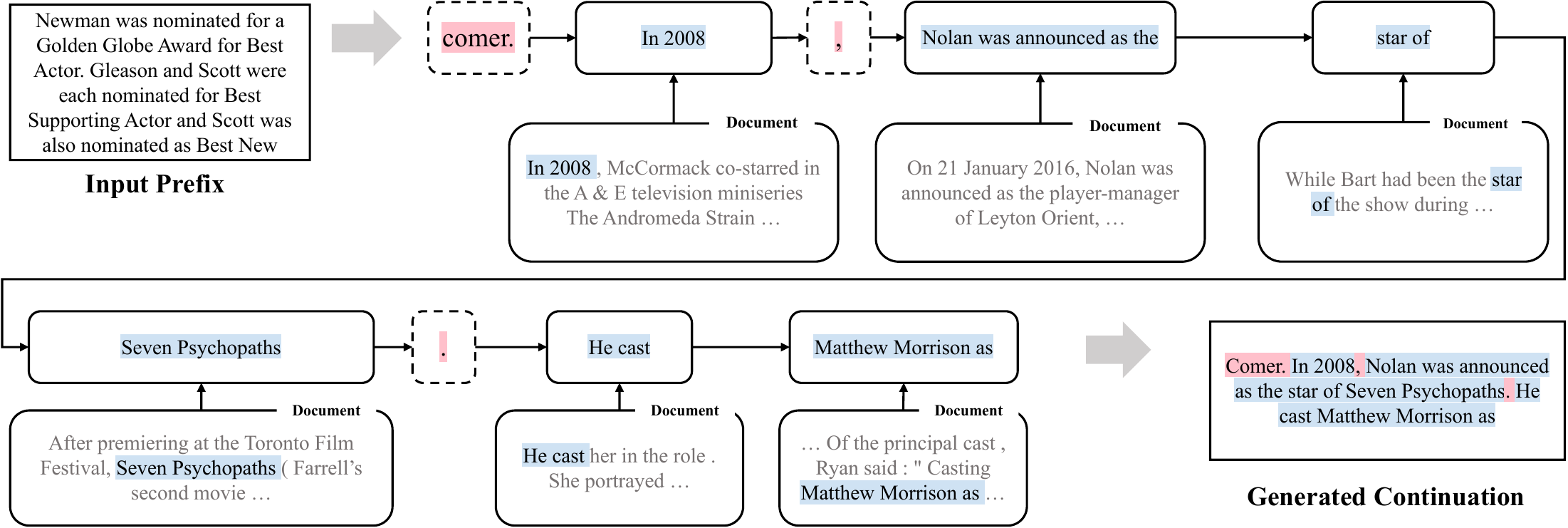}}
          \caption{An example generated by \ours. The dotted squares denote that the content (highlighted in \colorbox{pink}{red})is generated from the token vocabulary, and the solid squares denote that the content (highlighted in \colorbox{lightblue}{blue}) is copied from other documents.}
          \label{img:case_study_appendix_1}
    \end{figure*}

\begin{figure*}[h]     
          \center{\includegraphics[width=\textwidth] {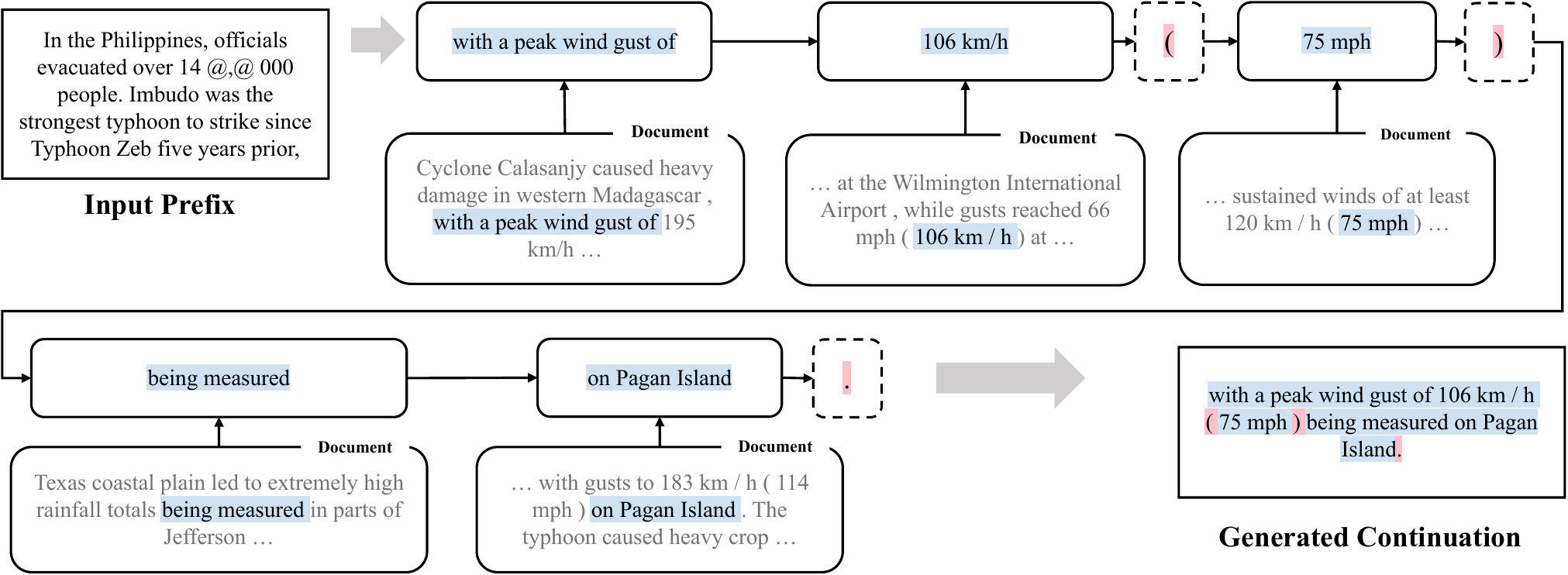}}
          \caption{An example generated by \ours. The dotted squares denote that the content (highlighted in \colorbox{pink}{red})is generated from the token vocabulary, and the solid squares denote that the content (highlighted in \colorbox{lightblue}{blue}) is copied from other documents.}
          \label{img:case_study_appendix_2}
    \end{figure*}

\begin{figure*}[h]     
          \center{\includegraphics[width=\textwidth] {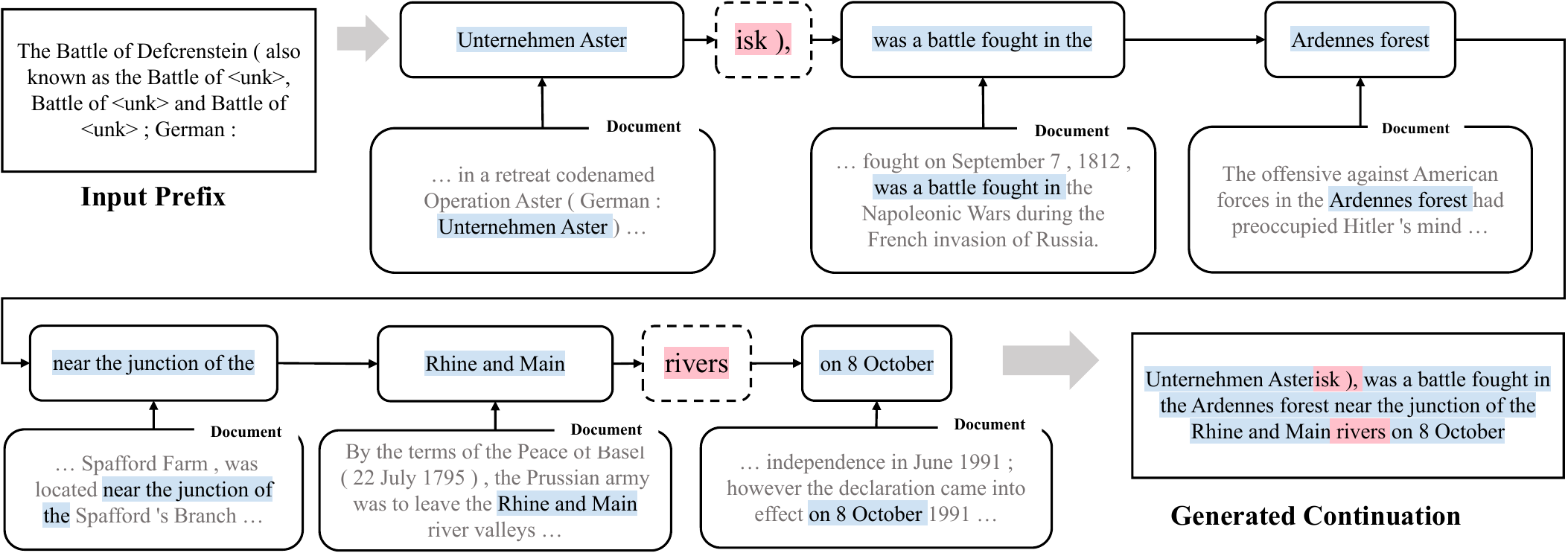}}
          \caption{An example generated by \ours. The dotted squares denote that the content (highlighted in \colorbox{pink}{red})is generated from the token vocabulary, and the solid squares denote that the content (highlighted in \colorbox{lightblue}{blue}) is copied from other documents.}
          \label{img:case_study_appendix_3}
    \end{figure*}

    In this section, we present some generated examples of \ours given some specific prefixes.
    As shown in Figure \ref{img:case_study_appendix_1}, \ref{img:case_study_appendix_2} and \ref{img:case_study_appendix_3}, it can be observed that the generated continuations are fluent and coherent with the given prefix.
    However, we also notice some flaws.
    For example, as shown in Figure \ref{img:case_study_appendix_2}, \ours copied the phrase \textit{75 mph} from the document \textit{... sustained winds of at least 120 km / h (75 mph)}, which is incoherent with the previous copied phrase \textit{106 km / h}.
    Moreover, as shown in Figure \ref{img:case_study_appendix_3}, \ours copied the phrase \textit{Rhine and Main} from the document (\textit{Bt the terms of the Peace of Basel (22 July 1795), the Prussian army was to leave the Rhine and Main river valleys ...}). However, the complete phrase should be \textit{Rhine and Main river valleys}, and \ours only copy a part of it, leading to inaccurate generation results (\textit{rivers}).
    %For example, in Figure \ref{img:case_study_appendix_2}, \ours lists some different types of matches in the game, such as \textit{elimination type matches}, \textit{tag team matches} and \textit{dark matches} from the documents that have the same topic (\textit{WWE video game}) with the prefix.

    %Additionally, we also show some bad cases generated by \ours. As shown in Figure \ref{img:case_study_4}, \ours retrieves a phrase \textit{Beavers)} that contains a right bracket. Since there is no left bracket in the prefix, this inappropriate phrase leads to not fluent and confusing generated continuation.
    %Moreover, as shown in Figure \ref{img:case_study_5}, in the fourth step, the retrieved phrase \textit{who become an important Norwegian poet} is incoherent for its prefix. In this case, the right bracket should be directly used to finish the text segments \textit{(The Bridge of the Vaaben)}.
    
    % \begin{figure*}[h]
    %       \center{\includegraphics[width=\textwidth] {img/case study 5.pdf}}
    %       \caption{An example generated by \ours. The dotted squares denote that the content (highlighted in \colorbox{pink}{red})is generated from the token vocabulary, and the solid squares denote that the content (highlighted in \colorbox{lightblue}{blue}) is copied from other documents.}
    %       \label{img:case_study_5}
    % \end{figure*}

\end{document}